\documentclass[12pt]{article}

\usepackage{amssymb,amsmath,amsfonts,latexsym,graphicx}
\usepackage[numbers]{natbib}
\usepackage{graphicx}
\usepackage{subcaption}
\usepackage{hyperref}

\begin{document}

\begin{center}
\Large \bf ClassifyViStA:WCE Classification with Visual understanding through Segmentation and Attention \rm

\vspace{1cm}


\large S. Balasubramanian$\,^a$, \large  Ammu Abhishek$\,^a$, \large  Yedu Krishna$\,^a$, \large   Darshan Gera$\,^a$

\vspace{0.5cm}

\normalsize


$^a$ Department of Mathematics and Computer Science, Sri Sathya Sai Institute of Higher Learning, Andhra Pradesh, India.

\vspace{5mm}


Email: {\tt sbalasubramanian@sssihl.edu.in}

\vspace{1cm}

\end{center}

\abstract{}
Gastrointestinal (GI) bleeding is a serious medical condition that presents significant diagnostic challenges, particularly in settings with limited access to healthcare resources. Wireless Capsule Endoscopy (WCE) has emerged as a powerful diagnostic tool for visualizing the GI tract, but it requires time-consuming manual analysis by experienced gastroenterologists, which is prone to human error and inefficient given the increasing number of patients. To address this challenge, we propose ClassifyViStA, an AI-based framework designed for the automated detection and classification of bleeding and non-bleeding frames from WCE videos. The model consists of a standard classification path, augmented by two specialized branches: an implicit attention branch and a segmentation branch. The attention branch focuses on the bleeding regions, while the segmentation branch generates accurate segmentation masks, which are used for classification and interpretability. The model is built upon an ensemble of ResNet18 and VGG16 architectures to enhance classification performance. For the bleeding region detection, we implement a Soft Non-Maximum Suppression (Soft NMS) approach with YOLOv8, which improves the handling of overlapping bounding boxes, resulting in more accurate and nuanced detections. The system’s interpretability is enhanced by using the segmentation masks to explain the classification results, offering insights into the decision-making process similar to the way a gastroenterologist identifies bleeding regions. Our approach not only automates the detection of GI bleeding but also provides an interpretable solution that can ease the burden on healthcare professionals and improve diagnostic efficiency. Our code is available at \href{https://github.com/1980x/WCEClassifyViStA}{ClassifyViStA}

\newpage
\section{Introduction}\label{sec1}
Gastrointestinal (GI) bleeding is a critical condition that encompasses a wide range of diseases, from acute to chronic conditions, and represents a major cause of morbidity and mortality worldwide. According to the World Health Organization (WHO), GI bleeding is responsible for approximately 300,000 deaths annually \cite{who2016global}. Early and accurate diagnosis of GI bleeding is essential for timely intervention, yet it remains a significant challenge, particularly in developing countries where there is a shortage of trained medical professionals \cite{patel2018challenges}. The conventional diagnostic approach using endoscopy, such as Wireless Capsule Endoscopy (WCE), requires gastroenterologist to manually analyze videos captured by the capsule, a process that can be time-consuming and prone to human error\cite{snyder2020wireless}.

\noindent WCE has proven to be a valuable tool for diagnosing GI bleeding, offering non-invasive and high-resolution imaging of the GI tract. However, the process of reviewing WCE videos is not only tedious but also inefficient, especially with the growing number of patients. On average, a gastroenterologist may take several hours to inspect the video frame-by-frame, which is a significant bottleneck in clinical practice \cite{hassan2019advances}. As a result, there is a need for innovative solutions that can assist in automating the detection of bleeding regions in WCE videos to reduce the workload on healthcare professionals and improve diagnostic accuracy. Recent advances in Artificial Intelligence (AI) and deep learning have shown promising potential in automating medical image analysis, including in the context of endoscopic images\cite{gao2020deep}. However, many existing models lack interpretability, making it difficult for healthcare professionals to trust and understand AI-driven decisions \cite{caruana2015modeling}. The Auto-WCEBleedGen challenge \cite{hub2024auto}, which aims to develop AI models for the automated classification and detection of bleeding and non-bleeding frames from WCE videos, provides an opportunity to address these issues and advance the field of AI-assisted diagnostics in GI bleeding.

\noindent In this paper, we introduce ClassifyViStA, a novel AI framework that combines classification, attention, and segmentation to automate the detection and classification of GI bleeding in WCE images. The framework is designed with two primary goals: (1) to improve classification performance by leveraging an ensemble of ResNet18 \cite{DBLP:journals/corr/HeZRS15} and VGG16 \cite{simonyan2014very} models, and (2) to provide interpretable predictions using segmentation masks that identify the bleeding regions in the images. Additionally, we enhance the detection of bleeding regions by applying Soft Non-Maximum Suppression (Soft NMS) to YOLOv8 \cite{10533619}, improving the accuracy of overlapping box handling. Our method provides a generalized AI solution that not only automates the diagnostic process but also ensures that predictions are interpretable, making it easier for healthcare professionals to trust the model's outputs.

\noindent By integrating these advances into a cohesive framework, ClassifyViStA aims to reduce the burden on gastroenterologists, enabling faster diagnoses and better patient outcomes. Our approach has the potential to transform GI bleeding diagnostics by providing a scalable and efficient tool for healthcare settings worldwide

\section{Methods}\label{sec2}

\begin{figure}[htbp]
    \centering
    \includegraphics[width=0.6\linewidth]{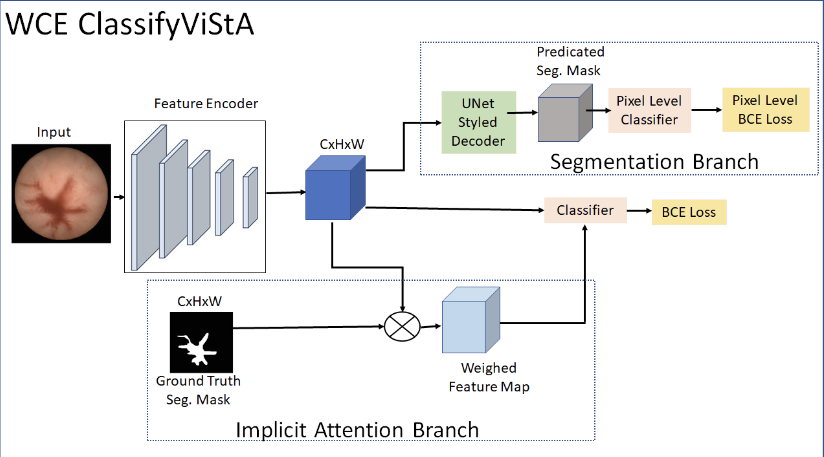}
    \caption{Architecture of the proposed model}
    \label{fig:enter-label}
\end{figure}

The proposed framework leverages a multi branch architecture with an ensemble-based classification backbone, augmented by implicit attention and segmentation branches to enhance performance and interpretability. From \autoref{fig:enter-label} it can be understood that the backbone of ClassifyViStA comprises two classification models, ResNet18 \cite{DBLP:journals/corr/HeZRS15} and VGG16 \cite{simonyan2014very}, working in an ensemble. Both models process the input WCE images independently, and their prediction probabilities are averaged to provide the final classification output. The use of two diverse architectures ensures complementary feature extraction, enhancing the robustness of the classification. The implicit attention branch incorporates ground truth segmentation masks to guide the classification process. Feature maps extracted from the encoder are weighted by these masks, emphasizing the regions of interest (ROIs) corresponding to bleeding areas. The weighted feature maps are then passed to the classification head, ensuring the network focuses primarily on bleeding regions for bleeding-class images. For non-bleeding images, the entire feature map is used, simulating global attention. This branch is active only during training, as segmentation masks are not available during inference. To complement classification and provide interpretability, a segmentation branch is incorporated. This branch utilizes a U-Net \cite{ronneberger2015u} style decoder to generate segmentation masks from encoder outputs. During training, the segmentation branch reconstructs ground-truth segmentation masks, enabling the network to learn spatial localization of bleeding regions. For nonbleeding images, the network generates zero-filled masks, ensuring consistent training dynamics.

\noindent The predicted segmentation masks serve the following purposes:

1. They act as explanations for classification decisions, mimicking the diagnostic reasoning of medical experts who focus on specific regions to identify bleeding.

2. They allow the identification of bright pixels for bleeding regions and dark pixels for nonbleeding regions, providing a natural and interpretable visualization of class predictions.

\noindent During inference, the implicit attention branch is omitted, since ground truth segmentation masks are unavailable. Class predictions are derived solely from the standard classification path. The segmentation branch generates predicted masks that are used for explanation purposes rather than to influence classification decisions.

\noindent Instead of relying on external explainability methods like LIME or SHAP, ClassifyViStA leverages its segmentation branch for intrinsic explainability. The predicted segmentation masks highlight the ROIs corresponding to bleeding regions, providing a visual explanation of the classification outcome. This aligns closely with the interpretive methods employed by medical professionals. For detection of bleeding regions, we integrate YOLOv8 with Soft Non-Max Suppression (Soft NMS). While YOLOv8 serves as a powerful detection framework, replacing standard NMS with Soft NMS during inference ensures a more refined handling of overlapping bounding boxes. This approach minimizes the risk of discarding partially correct detections, improving precision and recall on the validation set.

\noindent To increase the precision of classification, the final output of ClassifyViStA is derived by averaging the prediction probabilities of the ResNet18 and VGG16 models. This ensemble strategy harnesses the strengths of both architectures, mitigating biases inherent to individual models, and improving overall performance.

\section{Results}\label{sec3}

The dataset used for the model is \cite{hub2024auto}. For training we have used \cite{palakbleedingtrain} and for testing \cite{palakbleedingtest}.
\noindent The classification performance of the ClassifyViStA framework on the validation set is summarized in Table 1. The results demonstrate high accuracy, precision, recall, and F1-score, underscoring the robustness of the ensemble-based classification approach.

\begin{table}[h!]
\centering
\caption{Classification Performance on Validation Set}
\vspace{0.3cm} 
\begin{tabular}{|c|c|c|}
\hline
\textbf{S.No.} & \textbf{Metric} & \textbf{Value} \\
\hline
1 & Accuracy & 0.9962 \\
\hline
2 & Precision & 0.9962 \\
\hline
3 & Recall & 0.9962 \\
\hline
4 & F1-Score & 0.9962 \\
\hline
\end{tabular}
\end{table}

\begin{table}[h!]
\centering
\caption{Detection Performance on Validation Set}
\vspace{0.3cm} 
\begin{tabular}{|c|c|c|}
\hline
\textbf{S.No.} & \textbf{Metric} & \textbf{Value} \\
\hline
1 & Average Precision & 0.7715 \\
\hline
2 & MAP@0.5 & 0.726 \\
\hline
3 & MAP@0.5-0.95 & 0.483 \\
\hline
4 & Average IoU & 0.6405 \\
\hline
\end{tabular}
\end{table}

\noindent Table 2 presents the detection performance metrics on the validation set. The use of YOLOv8 with Soft NMS demonstrates effective bleeding region detection, achieving competitive Average Precision (AP), mean Average Precision (mAP), and Intersection over Union (IoU) scores.

\noindent The classification metrics highlight the efficacy of the ensemble strategy, achieving near-perfect values across all metrics. This confirms that combining ResNet18 and VGG16 effectively captures complementary features, enhancing overall classification performance.

\noindent The detection results indicate that incorporating Soft NMS in YOLOv8 improves the localization and identification of bleeding regions. The competitive Average Precision and IoU scores validate the model’s capability to detect and localize bleeding regions accurately, despite the challenges posed by the dataset.

\noindent Out of all our results we show few images that has diverse range of scenarios, including variations in illumination, anatomical regions (within the stomach), texture, and bounding box characteristics such as size (small, medium, and large) and significant overlap with the corresponding ground truth. Each selected image captures distinct challenges faced in real-world WCE analysis, ensuring a comprehensive evaluation of the detection model's robustness.

\begin{figure}[htbp]
    \centering
    \begin{subfigure}[b]{0.4\linewidth}
        \centering
        \includegraphics[width=\linewidth]{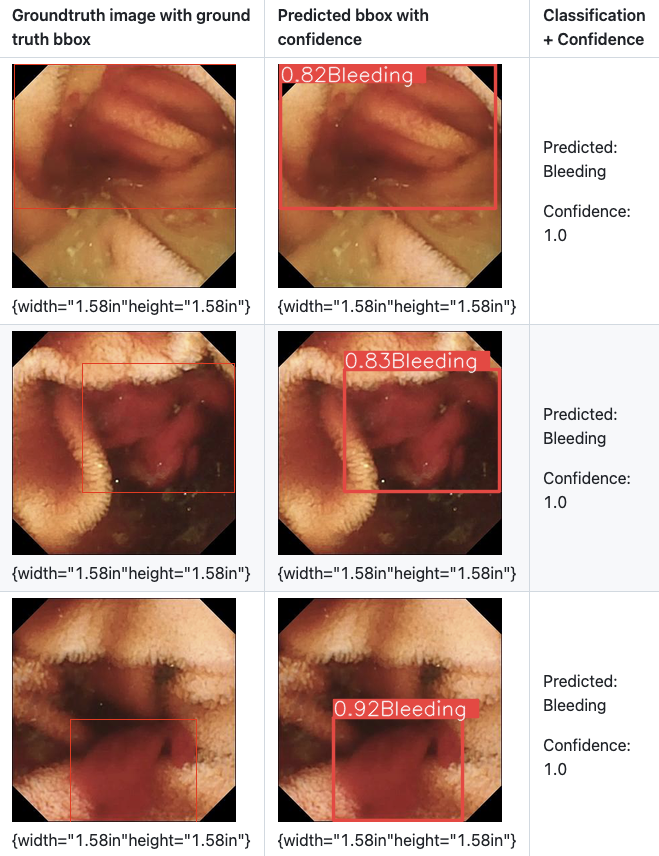}
        \caption{Classification and Detection with confidence and IOU detection}
        \label{fig:subfig1}
    \end{subfigure}
    \hfill
    \begin{subfigure}[b]{0.4\linewidth}
        \centering
        \includegraphics[width=\linewidth]{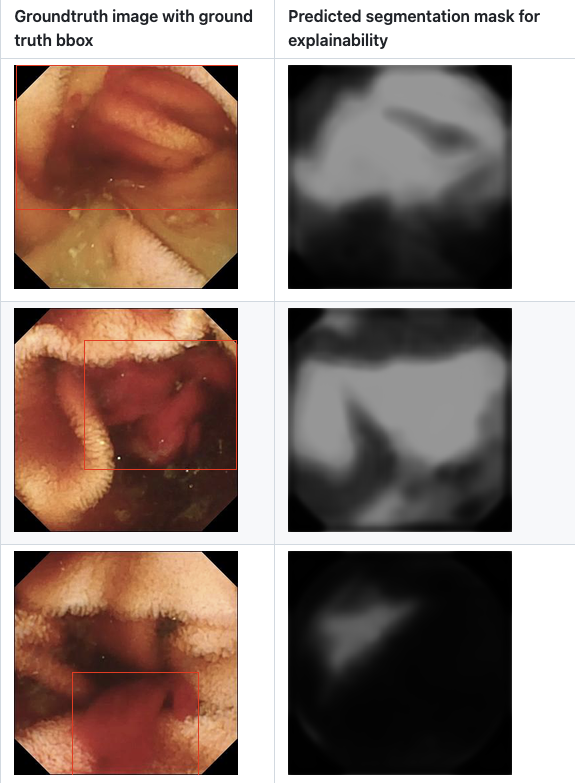}
        \caption{Interpretability Plot for Images}
        \label{fig:subfig2}
    \end{subfigure}
    \caption{Images of classification, detection, and interpretability plots.}
    \label{fig:side-by-side}
\end{figure}

\noindent For each image, the predicted bounding boxes and the ground truth detections are visually depicted, enabling a straightforward comparison of the model's performance. This qualitative analysis complements the quantitative metrics, highlighting the model's ability to adapt to diverse visual features and effectively localize bleeding regions under varying conditions. The inclusion of edge cases and challenging scenarios in this visual evaluation provides deeper insights into the strengths and limitations of the proposed framework.

\begin{figure}[htbp]
    \centering
    \begin{minipage}[b]{0.45\linewidth}
        \centering
        \includegraphics[width= 0.7\linewidth]{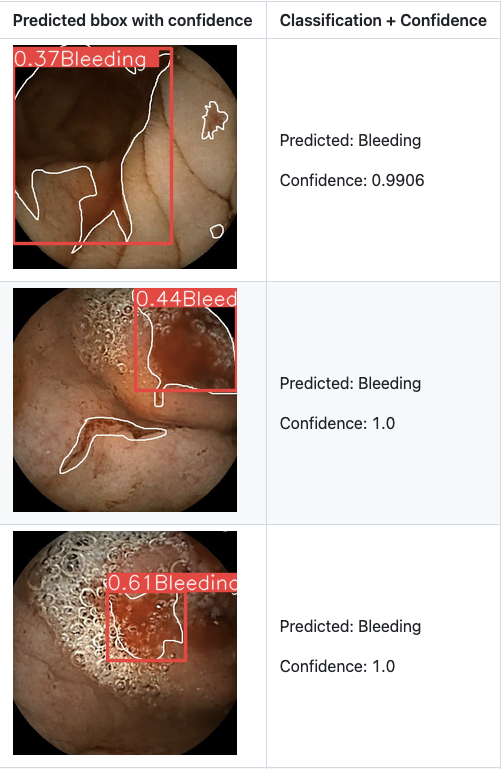}
        \caption{Classification and Detection from testset 1}
        \label{fig:image1}
    \end{minipage}
    \hspace{0.1cm}
    \begin{minipage}[b]{0.45\linewidth}
        \centering
        \includegraphics[width=0.7\linewidth]{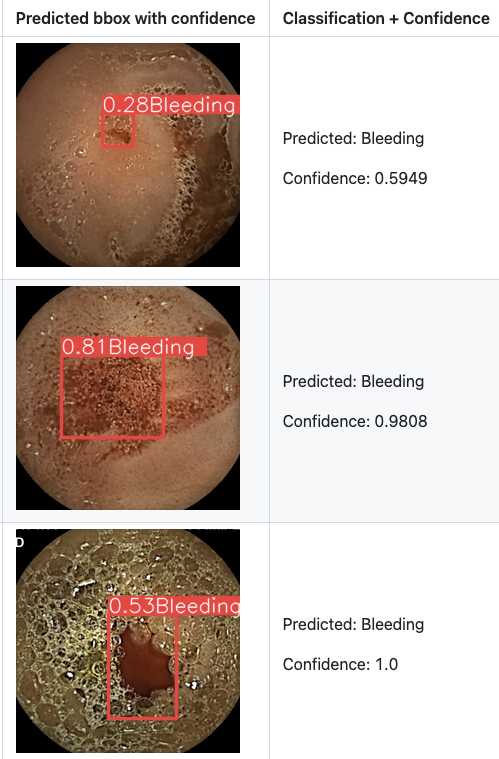}
        \caption{Classification and Detection from testset 2}
        \label{fig:image2}
    \end{minipage}

    \vspace{0.5cm} 

    \begin{minipage}[b]{0.45\linewidth}
        \centering
        \includegraphics[width=0.7\linewidth]{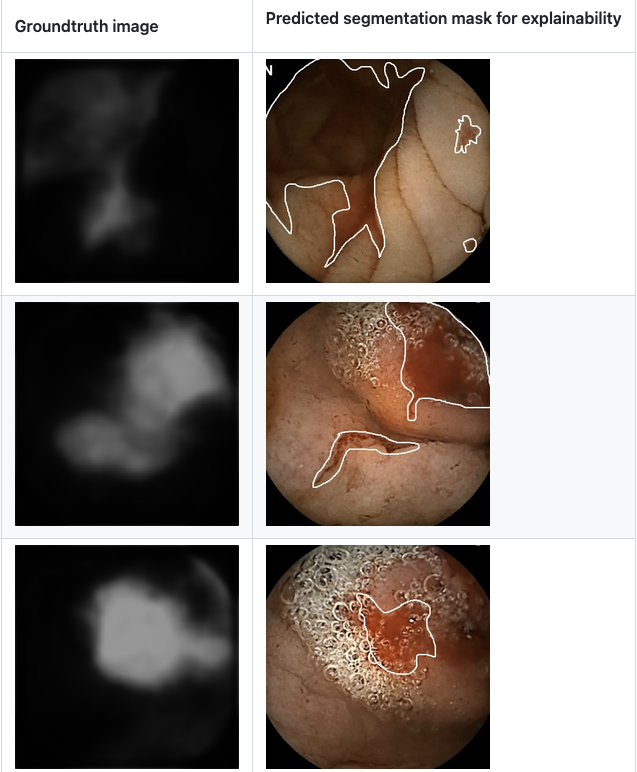}
        \caption{Interpretability Plots from testset 1}
        \label{fig:image3}
    \end{minipage}
    \hspace{0.1cm}
    \begin{minipage}[b]{0.45\linewidth}
        \centering
        \includegraphics[width=0.7\linewidth]{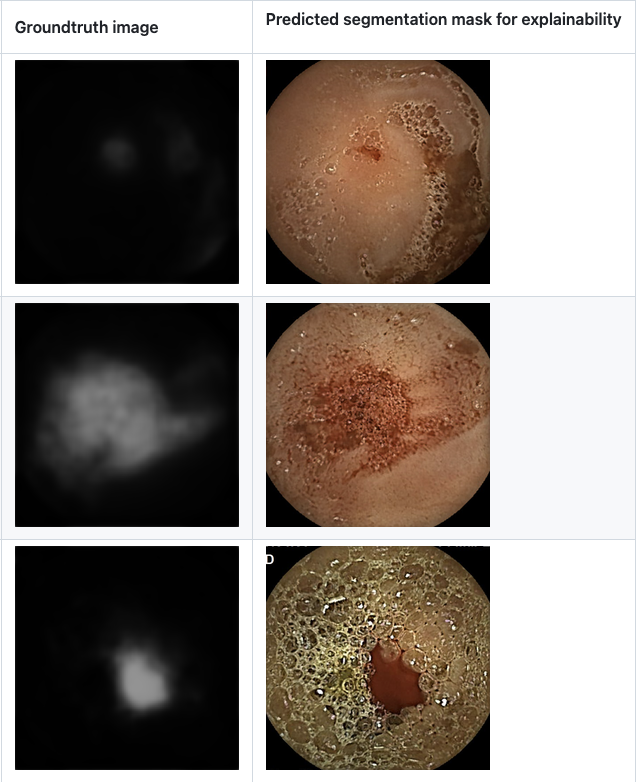}
        \caption{Interpretability Plots from testset 2}
        \label{fig:image4}
    \end{minipage}
    
    \label{fig:2x2-matrix}
\end{figure}

\newpage
\section{Discussion}\label{sec4}
During the development and evaluation of ClassifyViStA, we made several key observations that highlight areas for improvement and potential future enhancements. Firstly, we observed varying accuracies across different validation sets, with some results significantly outperforming the quoted metrics. This variability underscores the importance of having a fixed or hold-out validation set to ensure consistent benchmarking. While we maintained a fixed random state of 42 throughout the experiments to ensure reproducibility, a standardized validation strategy could further enhance the reliability of reported results.

\noindent Secondly, the pre-processing of Test Dataset 1 \cite{palakbleedingtest} posed an additional challenge. The provided images contained superimposed boundaries, which required preprocessing to achieve optimal performance. It would have been more effective if these boundaries had been provided as separate segmentation masks, allowing for direct integration into the segmentation pipeline. This adjustment would streamline pre-processing efforts and potentially improve model performance by eliminating unnecessary pre-processing complexities.

\noindent These observations highlight the importance of dataset standardization and preprocessing efficiency in achieving robust and reproducible results, and they provide valuable insights for future iterations of this framework.

\section{Conclusion}\label{sec5}

In conclusion, ClassifyViStA offers a promising solution for automating the detection and classification of GI bleeding in WCE videos. By integrating attention mechanisms and segmentation for interpretability, and using ensemble models for improved accuracy, the framework significantly enhances diagnostic efficiency. This approach not only supports gastroenterologists in their clinical decision-making but also addresses the challenges posed by the growing patient load in healthcare settings. The results highlight the potential of AI in transforming the diagnostic process for GI bleeding, providing a more scalable and effective tool for medical professionals.

\section{Acknowledgments}\label{sec6}
This work is dedicated to Bhagawan Sri Sathya Sai Baba, the Founder Chancellor of Sri Sathya Sai Institute of Higher Learning. As participants in the Auto-WCEBleedGen Version V1 Challenge, we fully comply with the competition rules as outlined in \cite{hub2024auto} and the challenge website. Our methods have used the training and test data sets provided in the official release in \cite{palakbleedingtrain} and \cite{palakbleedingtest} to report the results of the challenge.

\bibliographystyle{unsrtnat}
\bibliography{main}

\end{document}